\pdfoutput=1

\documentclass[11pt]{article}

\usepackage{emnlp2021}

\usepackage{times}
\usepackage{latexsym}

\usepackage[T1]{fontenc}
\usepackage[utf8]{inputenc}

\usepackage{microtype}

\usepackage{array}
\newcolumntype{L}{>{\centering\arraybackslash}m{1.2cm}}

\usepackage{multirow}

\title{Machine Translation of Low-Resource Indo-European Languages} %

\author{Wei-Rui Chen ~~~~~~~~  Muhammad Abdul-Mageed \\
\normalsize Natural Language Processing Lab  \\
  \normalsize The University of British Columbia\\
      
  \texttt{ \small \{weirui.chen,muhammad.mageed\}@ubc.ca}
  }

\begin{document}
\maketitle
\begin{abstract}
In this work, we investigate methods for the challenging task of translating between low-resource language pairs that exhibit some level of similarity. In particular, we consider the utility of transfer learning for translating between several Indo-European low-resource languages from the Germanic and Romance language families. In particular, we build two main classes of transfer-based systems to study how relatedness can benefit the translation performance. The \textit{primary} system fine-tunes a model pre-trained on a \textit{related} language pair and the \textit{contrastive} system fine-tunes one pre-trained on an \textit{unrelated} language pair. Our experiments show that although relatedness is not necessary for transfer learning to work, it does benefit model performance.

\end{abstract}

\section{Introduction}\label{sec:introduction}

Machine translation (MT) is currently one of the hot application areas of deep learning, with neural machine translation (NMT) achieving outstanding performance where large amounts of parallel data are available~\cite{koehn-knowles-2017-six,ranathunga2021neural, luong2015effective, DBLP:journals/corr/abs-1708-09803}. In low-resource settings, transfer learning methods have proven useful for improving system performance~\cite{zoph2016transfer,DBLP:journals/corr/abs-1708-09803,DBLP:journals/corr/abs-1809-00357}. In this work, we focus on studying NMT, including in the low-resource scenario. In particular, we focus our attention on investigating the effect of \textit{\textbf{language relatedness}} on the transfer process. We define relatedness of a pair of languages based on belonging to the same language family. That is, by `related' we mean `within the \textit{same} language family' whereas by `unrelated' we mean `belong to two \textit{different} language families'. For example, we call English and Swedish related since they belong to the Germanic language family but English and French \textit{not} related since the latter belongs to the Romance language family. As an analogy to human learning, we would like to ask: \textit{if there are two translators (pre-trained models), for example one Catalan $\rightarrow$ Spanish translator and one Catalan $\rightarrow$ English translator, will they (after extra training, i.e., fine-tuning/transfer learning process) have different abilities to translate from Catalan into Occitan?} If the Catalan-Spanish translator proves to perform better Catalan $\rightarrow$ Occitan, we may attribute this to Spanish and Occitan being members of the Romance language family while English being a member of the, Germanic, different family.

Of particular interest to us are two sets of languages belonging to two different language families, one set to Romance and the other set to Germanic. For the former set, we take Catalan (ca), Italian (it), Occitan (oc), Romanian (ro), and Spanish (es); and we take English (en) for the latter set. We note that both Romance and Germanic are two branches of the larger Indo-European language family, and hence there are some level of relatedness between all the languages we study in this work. Nevertheless, languages in Romance and Germanic differ in some syntactic structures. For example, the position of attributive adjectives in Germanic languages is before the noun while it is after the noun for Romance languages~\cite{van2014adjective}. Despite differences, the writing system of all languages in this work is the Latin script. This can be beneficial to transfer learning because these languages can potentially share common lexical items or morphemes, which may facilitate the transfer learning process.

\begin{table}
\centering
\begin{tabular}{ccc}
\hline
\textbf{Lang. Pair} & \textbf{Primary} & \textbf{Contrastive}\\
\hline
ca-it & ca-it/ca-es & ca-en  \\
ca-oc & ca-es & ca-en  \\
ca-ro & ca-es & ca-en  \\
\hline
\end{tabular}

\caption{Pre-trained model choices for our primary and contrastive NMT systems. }
\label{tab:model_choices}
\end{table}

As mentioned, we adopt transfer learning since it has been shown to improve translation quality for low-resource language pairs. For example,~\citet{zoph2016transfer} sequentially build a \textit{parent} model and a \textit{child} model where each is trained, respectively, on high-resource and low-resource language pairs with the child model retaining parameters from the parent model. In  addition,~\citet{DBLP:journals/corr/abs-1708-09803} successfully transfer knowledge from a parent model to a child model where both models are trained on low-resource but related language pairs.~\citet{DBLP:journals/corr/abs-1809-00357} also adopt a similar approach to these previous works, but based their work on the Transformer architecture~\cite{NIPS2017_3f5ee243} instead of using a recurrent encoder-decoder network with attention ~\cite{bahdanau2015neural,luong-etal-2015-effective}.

Our work builds on these studies. Namely, we empirically apply transfer learning under different conditions to members of various language families. The outcomes of our work are similar to those of \citet{zoph2016transfer,DBLP:journals/corr/abs-1809-00357}. That is, while we find relatedness to be beneficial, a positive transfer between an unrelated language pair can still be possible (although with a potentially diminished performance).

The rest of this paper is organized as follows: In Section~\ref{sec:lit}, we overview related work. We introduce our datasets and experiments in Section~\ref{sec:exp}. In Section~\ref{sec:res}, we present and analyze our results. We conclude in Section~\ref{sec:conc}.

\section{Background}\label{sec:lit}

\subsection{Transfer learning}
Transfer learning is a machine learning approach that aims at transferring the knowledge of one task to another. As an analogy to human learning, one who masters the skills to ride a bicycle may transfer the knowledge to riding a motorcycle because these two tasks share common abilities such as maintaining balance on a two-wheel moving vehicle~\cite{pan2009survey,weiss2016survey}. We employ transfer learning to port knowledge from a model trained on one pair of languages to another. We now discuss transfer learning in NMT.

\subsection{Transfer learning in Machine Translation}~\citet{zoph2016transfer} design a framework where a parent model is trained on a high-resource language pair while retaining model parameters for the child model to start fine-tuning with. Using this method,~\citet{zoph2016transfer} improve system performance by an average of $5.6$ BLEU points. The improvement is realized by transferring what is learnt in the high-resource language pair to the low-resource language pair. The Uzbek-English model obtains $10.7$ BLEU score without the parent model and improves to $15.0$ with the French-English parent model. The Spanish-English model has $16.4$ BLEU score without the parent model and $31.0$ with the French-English parent model. These results show that applying transfer learning contributes $4.3$ and $14.6$ BLEU points gain. Based on results from ~\citet{zoph2016transfer}, the closer the two source languages, the more performance gain acquired. Due to the relatedness between Spanish and French (both are members of the Roman language family), performance gain is higher for this pair.

Following previous work,~\citet{DBLP:journals/corr/abs-1708-09803} design a paradigm similar to that of~\citet{zoph2016transfer} but maintain one major difference. In particular,~\citet{DBLP:journals/corr/abs-1708-09803} try to make use of relatedness between the parent and child models at the vocabulary level: instead of randomly mapping tokens in the parent and child vocabulary, they retain the parent tokens for the child model if these tokens exist in child language pair. This approach is based on two assumptions - (i) the lexicons of the parent and child language pair have at least some partial overlap and (ii) these identical tokens have similar meaning. Instead of the word-level tokenization in~\citet{zoph2016transfer}, ~\citet{DBLP:journals/corr/abs-1708-09803} use Byte Pair Encoding (BPE)~\cite{gage1994new, sennrich-etal-2016-neural} to obtain subword tokens which may increase the number of overlapped tokens between the parent and child models. Improvement of $0.8$ and $4.3$ in BLEU score were obtained for the Turkish-English and Uyghur-English child models as transferred from an Uzbek-English parent model.

Following the previous two works, ~\citet{DBLP:journals/corr/abs-1809-00357} take a similar approach but use the Transformer architecture. They obtain an improvement of $3.38$ BLEU for an English-Estonian child model transferred from an English-Czech parent model. Similarly, \citet{neubig-hu-2018-rapid} add a second language related to the added low-resource language to avoid overfitting when fine-tuning. This mechanism has shown to be effective. Other works have investigated NMT approaches to similar languages by pre-training new language models on the low-resource languages~\citep{nagoudi:2021:indt5} or without necessarily applying transfer learning~\citep{przystupa2019neural,adebara2020translating, barrault2019findings, Barrault2020FindingsOT}, and there are several works on low resource languages~\citep{adebara2021translating}. We now introduce our experimental settings.
\section{Experimental Settings}\label{sec:exp}
\begin{table*}[h]
\centering
\begin{tabular}{Lc|cc|rrrrr}
\hline
& & \multicolumn{2}{c|}{\textbf{Baseline}} & \multicolumn{5}{c}{\textbf{Our models}} \\
\hline
\textbf{Pre-trained Model} & \textbf{Lang. Pairs} & \textbf{BLEU} & \textbf{chrF} & \textbf{BLEU} & \textbf{chrF} & \textbf{TER} & \textbf{COMET} & \textbf{BertScore}\\
\hline
ca-it & ca-it & 29.31 & 0.583 & \textbf{35.06} & \textbf{0.622} & 0.477 & 0.391 & 0.886 \\
ca-es & ca-it & 7.07 & 0.370 & \textbf{33.13} & \textbf{0.602} & 0.499 & - & - \\
ca-es & ca-oc & 12.56 & 0.472 & \textbf{59.93} & \textbf{0.787} & 0.254 & 0.538 & 0.928 \\
ca-es & ca-ro & 4.43 & 0.266 & \textbf{11.24} & \textbf{0.354} & 0.855 & -0.908 & 0.749 \\
\hline

\end{tabular}
\caption{Primary system results. We did not submit the ca-it language pair fine-tuned on the ca-es pre-trained model to the WMT2021 shared task, and hence the results are calculated by ourselves with Sacrebleu.}
\label{tab:primary_reuslts}
\end{table*}

\begin{table*}
\centering
\begin{tabular}{Lc|cc|rrrrr}
\hline
& & \multicolumn{2}{c|}{\textbf{Baseline}} & \multicolumn{5}{c}{\textbf{Our models}} \\
\hline
\textbf{Pre-trained Model} & \textbf{Lang. Pairs} & \textbf{BLEU} & \textbf{chrF} & \textbf{BLEU} & \textbf{chrF} & \textbf{TER} & \textbf{COMET} & \textbf{BertScore}\\
\hline
ca-en & ca-it & 1.97 & 0.249 & \textbf{25.46} & \textbf{0.539} & 0.574  & -0.263 & 0.844 \\
ca-en & ca-oc & 2.16 & 0.258 & \textbf{51.46} & \textbf{0.736} & 0.316 & 0.259 & 0.905 \\
ca-en & ca-ro & 1.59 & 0.209 & \textbf{8.61} & \textbf{0.311} & 0.884 & -1.119 & 0.725\\
\hline

\end{tabular}
\caption{Contrastive system reuslts}
\label{tab:contrastive_reuslts}
\end{table*}
\subsection{Languages \& Settings}
We carry out experiments on three language pairs: \textit{ca-it}, \textit{ca-oc}, and \textit{ca-ro}. The number of parallel sentences of each dataset is shown in Table~\ref{tab:dataset_distribution}. 
Training data are from OPUS~\cite{tiedemann-2012-parallel}, particularly version 1 of the WikiMatrix datasets~\cite{schwenk-etal-2021-wikimatrix}. They are data of child language pair and are used to fine-tune pre-trained model. Development and test data are provided by organizers of the Multilingual Low-Resource Translation for Indo-European Languages shared task. The shared task is hosted in EMNLP 2021 Sixth Conference on Machine Translation (WMT21).

\begin{table}
\centering
\begin{tabular}{crrr}
\hline
\textbf{Lang. Pairs} & \textbf{Train} & \textbf{Dev} & \textbf{Test}\\
\hline
ca-it & 1,143,531 & 1269 & 1743 \\
ca-oc & 138,743 & 1269 & 1743 \\
ca-ro & 490,064 & 1269 & 1743 \\
\hline
\end{tabular}

\caption{Distribution of dataset}
\label{tab:dataset_distribution}
\end{table}

We build two systems: a \textit{\textbf{Primary system}} and a \textit{\textbf{Contrastive system}}. The primary system fine-tunes pre-trained \textit{ca-es} and \textit{ca-it} models, while the contrastive system fine-tunes a pre-trained \textit{ca-en} model as shown in Table~\ref{tab:model_choices}. The primary and contrastive systems serve as context for studying the role of language \textit{\textbf{relatedness}} in transfer learning in NMT. We submitted predictions of the two systems to the WMT2021 shared task, and evaluation was based on blind test sets with a number of metrics as run by shared task organizers. An exception is the \textit{ca-it} language pair fine-tuned on top of the \textit{ca-es} pre-trained model--we train the model of this pair post shared task formal evaluation.

\subsection{Model Architecture} \label{subsec:experiment_settings_models}
We leverage publicly accessible pre-trained models on Huggingface~\cite{wolf-etal-2020-transformers} from Helsinki-NLP~\cite{TiedemannThottingal:EAMT2020}. The pre-trained MT models released by Helsinki-NLP are trained on OPUS. These models are Transformer-based implemented in the Marian-NMT framework~\cite{junczys-dowmunt-etal-2018-marian}. Each model has six self-attention layers in both the encoder and decoder sides, and each layer has eight attention heads. The tokenization method is SentencePiece~\cite{kudo-richardson-2018-sentencepiece} which produces vocabulary of size $49,621$, $21,528$ and $55,255$ for \textit{ca-es, ca-it}, and \textit{ca-en}  models, respectively.

\subsection{Approach}
 The pre-trained models are chosen based on the degree of \textit{relatedness} of the original target language on which the model is trained and the new target language on which the model is fine-tuned. \textit{Primary system} takes related languages while contrastive system takes unrelated languages. Since Catalan, Italian, Occitan, Romanian, and Spanish are all members of the Roman language family, we take \textbf{ca-es} as our pre-trained MT model for transfer learning. As English is a member of the Germanic language family, we use a \textbf{ca-en} pre-trained model for our transfer learning. Our model choices are summarized in Table~\ref{tab:model_choices}.

Without modifying the architecture of the MT pre-trained models, all architecture-related hyperparameters are identical to the original edition. As for hyperparameters related to fine-tuning, the number of beams for beam search is modified from four for pre-training to six for fine-tuning. The batch size is set to be $25$. Pre-trained models are further fine-tuned for $30,000$ steps on OPUS bitext. The checkpoint with the lowest validation loss is then selected as our best model for prediction. 

Similar to \citet{zoph2016transfer, DBLP:journals/corr/abs-1708-09803, DBLP:journals/corr/abs-1809-00357}, to achieve transfer learning, we retain the parameters of the parent model when fine-tuning the child model. Besides, parent and child models share a common vocabulary. That is, we do not build distinct vocabularies for the parent model and child models. A shared vocabulary can contribute to better transfer learning since all our language pairs employ the same Latin writing system. We suspect a shared vocabulary is more influential when the two languages are related to each other since the languages may have common morphemes, lexical items, or syntactical structure. For unrelated languages, a shared vocabulary may not hurt since the token embeddings are not frozen throughout the fine-tuning process. That is, token embeddings can still be updated to attain better representations during training.

\subsection{Baseline Models}

To demonstrate the effectiveness of our transfer learning approach, we provide a baseline model for each language pair that is simply a parent model (a pre-trained model) without any fine-tuning on data of the child language pair. 

\subsection{Evaluation}
The adopted metrics are BLEU~\cite{papineni-etal-2002-bleu}, chrF~\cite{popovic-2015-chrf}, TER~\cite{olive2005global}, COMET~\cite{rei-etal-2020-comet}, and BERTScore~\cite{zhang2019bertscore}. BLEU, chrF and TER are measured with the implementation of \textit{Sacrebleu}~\cite{post-2018-call}\footnote{https://github.com/mjpost/sacrebleu}.

\section{Results and Analysis}\label{sec:res}
\subsection{Primary and Contrastive Systems}

As can be seen in the rightmost five columns in Table~\ref{tab:primary_reuslts} and Table~\ref{tab:contrastive_reuslts}, primary system outperforms contrastive system across all metrics. We believe that \textbf{ca-es} pre-trained MT model performs better transfer learning because Spanish is closer to Italian, Occitan, and Romanian than English is to these languages. These results, as such, indicate that transfer learning between related language pairs can produce better performance than between unrelated language pairs.

\subsection{Baseline and Fine-tuned Models}
Our results in Table~\ref{tab:primary_reuslts} and Table~\ref{tab:contrastive_reuslts} show the effectiveness of transfer learning for both related and \textit{unrelated} language pairs. This is the case since both systems experience a performance gain after fine-tuning.

As an interesting observation, it seems counter-intuitive to have the unrelated language pairs experience slightly higher performance gain. For example, regarding \textbf{ca-oc} language pair, the transfer learning provides $47.37$ BLEU score improvement transferring from \textbf{ca-es} parent model but $49.3$ BLEU score improvement transferring from \textbf{ca-en} parent model. We suspect this is because in our work, when fine-tuning, we fix source language and alter the target language.

Unlike multilingual MT models which requires target language label to be prepended at the beginning of a source sentence~\cite{johnson-etal-2017-googles} or notifying the model what target language is for this forward propagation~\cite{liu-etal-2020-multilingual-denoising}, the pre-trained models we use in this work are bilingual models which lack a mechanism to provide the model any information about current target language. Therefore, the \textbf{ca-en} pre-trained model does not know it should now be translating Catalan to Occitan instead of English. Due to producing prediction in an incorrect target language, the metrics will be very poor. After fine-tuning the parent models on data of the child language pairs, the models are likely abler to produce prediction in the correct target language. Due to baseline metrics being too low, the difference in metric values between non-fine-tuned (baseline) and fine-tuned models are large and that is why the performance gain can be higher in contrastive system than in primary system.

\section{Conclusion}\label{sec:conc}
In this work, we confirm previous works showing that transfer learning benefits NMT. Besides, an empirical comparison between transferring from related and unrelated languages shows that relatedness is not strictly required for knowledge transfer, but it does result in higher performance than transferring with unrelated languages.

\section*{Acknowledgements}
We appreciate the support from the Natural Sciences and Engineering Research Council of Canada, the Social Sciences and Humanities Research Council of Canada, Canadian Foundation for Innovation, Compute Canada (\url{www.computecanada.ca}), and UBC ARC--Sockeye (\url{https://doi.org/10.14288/SOCKEYE}).

\bibliography{anthology,custom,some-refs}
\bibliographystyle{acl_natbib}

\end{document}